\documentclass{article}

\usepackage[final, nonatbib]{neurips_2019}

\usepackage[utf8]{inputenc} 
\usepackage[T1]{fontenc}    
\usepackage{hyperref}       
\usepackage{url}            
\usepackage{booktabs}       
\usepackage{amsfonts}       
\usepackage{nicefrac}       
\usepackage{microtype}      

\title{Auto-Rotating Perceptrons}

\author{
  Daniel Saromo, Elizabeth Villota and Edwin Villanueva \\
  Pontificia Universidad Católica del Perú\\
  \texttt{daniel.saromo@pucp.pe}, \texttt{\{evillota,ervillanueva\}@pucp.edu.pe} \\
}

\usepackage{cite}

\newcommand{\blaTeoName}{definition}
\usepackage{amsthm} 
\theoremstyle{\blaTeoName}
\newtheorem{\blaTeoName}{Theorem} 

\theoremstyle{remark}

\usepackage{amsmath}

\usepackage{amssymb}

\usepackage{svg}

\begin{document}

\maketitle

\section{Abstract}

This paper proposes an improved design of the perceptron unit to mitigate the vanishing gradient problem. This nuisance appears when training deep multilayer perceptron networks with bounded activation functions. The new neuron design, named auto-rotating perceptron (ARP), has a mechanism to ensure that the node always operates in the dynamic region of the activation function, by avoiding saturation of the perceptron. The proposed method does not change the inference structure learned at each neuron. We test the effect of using ARP units in some network architectures which use the sigmoid activation function. The results support our hypothesis that neural networks with ARP units can achieve better learning performance than equivalent models with classic perceptrons.

\section{Introduction}


Deep neural networks (DNN) models are widely used for inference problems in several areas, such as object detection \cite{han2018advanced}, pattern recognition \cite{zhang2018drawing} and image reconstruction \cite{rojas2017learning}.
Theoretically, the stacking architecture of these models allows them to learn any mapping function from input to output variables   \cite{bengio2009learningBook, lecun2015deep, schmidhuber2015deep}. However, in practice, DNN models are very difficult to train when using neural units with bounded activation functions \cite{he2016deepRESNET}. In such cases, the computation of the error gradients, needed by most learning methods to adjust the network weights, becomes problematic due to the exponential decrease of the gradients as we go down to the initial layers, which can cause slow learning, premature convergence and poor model inference performance \cite{nielsen2015neural, xu2016revise}. 
This issue is known as the vanishing gradient problem (VGP) \cite{bengio1994learning, hochreiter2001gradient}. Researchers have proposed a plethora of unbounded activation functions (e.g. ReLU, PReLU, leaky ReLU, ELU, etc; a review can be found in \cite{nwankpa2018activation}) as a way to overcome the VGP. Nonetheless, some authors have proposed bounded versions of such activation functions that show to be effective in alleviating the training instability of DNN models \cite{liew2016bounded}. 


In this paper, we propose a different approach to tackle the VGP in DNN learning. Instead of worrying about the activation function, we design a mechanism for the pre-activation phase. The intuition of this mechanism, called auto-rotation (AR),  is to avoid the activation function derivative to take small values (i.e., maintaining the operation of the neural unit in its dynamic region). We show the advantage of
this mechanism when applied to multilayer perceptron networks (MLP) by improving their learning performance.

\section{Auto-Rotating Perceptron (ARP)}

Recalling, a perceptron unit is a function that maps an input vector $\mathbf{x} \in \mathbb{R}^{n}$ to an output $\hat{y}\in \mathbb{R}$. This mapping is done in two steps. First, a weighted sum of the inputs is computed: $f(\mathbf{x})=\mathbf{w} \cdot \mathbf{x} + w_0 \, x_0$, where $\mathbf{w} \in \mathbb{R}^{n}$ is the weight vector and $x_0=1$. Then, a non-linear function $\sigma(\cdot)$ (i.e., the activation function) is applied to the weighted sum to obtain the neuron output: $\hat{y} = \sigma(f(\mathbf{x}))$.      


We define the $n$-dimensional hyperplane $\varphi$ in an ${(n+1)}$-dimensional orthogonal space as the points whose first $n$ dimensions are the input vector coordinates and the additional dimension being $f(\mathbf{x})$. In other words, we create the surface $\varphi  = \langle \mathbf{x}, f(\mathbf{x}) \rangle \subset \mathbb{R}^{n+1}$. When taking the points of $\varphi$ where $f(\cdot)=0$, we generate the boundary $\Gamma \subset \mathbb{R}^{n}$ that separates the input space into two regions with a different sign for its $f(\cdot)$ value (i.e., the region with $f(\mathbf{x})>0$ and the other one with $f(\mathbf{x})<0$).

When the neural units are  arranged into hierarchical structures, which occurs in an MLP network, the perceptrons of each hidden layer learn an abstract feature from the information given by the previous level. We can interpret that the regions defined by the boundary $\Gamma = \langle \mathbf{x}, 0\rangle \cap \varphi$ capture the presence or absence of an abstract inferred feature. 
What is essentially done in the training stage is to learn the non-discrete surface $\varphi$, whose intersection with the input space (i.e., the set of points of $\varphi$ where $f(\mathbf{x})=0$) is the boundary $\Gamma$ that defines the feature extracting capability of the unit. 
We propose to modify the inclination of the surface $\varphi$ without changing the boundary $\Gamma$. This is done by rotating $\varphi$ using the boundary  $\Gamma$  as the rotation axis. 


\label{theorem_HR}
  This makes sense, since with a rotation of $\varphi$ we can change the weighted input that goes into the activation function, but at the same time keeping the boundary $\Gamma$ unchanged. 
  Mathematically, we obtained that a rotation of the perceptron's $n$-dimensional hyperplane $\varphi  = \langle \mathbf{x}, f(\mathbf{x}) \rangle \subset \mathbb{R}^{n+1}$, around an $(n-1)$-dimensional axis $\Gamma = \langle \mathbf{x}, 0\rangle \subset \mathbb{R}^{n}$ (where $\Gamma \subset \varphi$), can be achieved by multiplying a real scalar value $\rho$ to all the weights $w_i \in \mathbf{w}$ and the bias $w_0$. The new $n$-dimensional hyperplane $\hat{\varphi} = \langle \mathbf{x}, g(\mathbf{x}) \rangle  \subset \mathbb{R}^{n+1}$ is defined in terms of the weighted sum $g(\mathbf{x})= 
    \rho \, \mathbf{w} \cdot \mathbf{x} + \rho\,w_0$, the rotation coefficient $\rho = \frac{L}{\left| f(\mathbf{x}_Q)  \right|}$ and the hyperparameters $L \in \mathbb{R}$ (which defines the limits of the activation function dynamic range) and $\mathbf{x}_Q \in \mathbb{R}^{n}$ (which is a point in $\mathbb{R}^{n}$ outside the input data range).
  
Notice that the rotation mechanism acts independently in each neural unit, adjusting dynamically the pre-activation phase (since $\rho$ depends on the perceptron weights) to prevent the activation function to saturate. Because of this auto-adaptation, we named the new unit auto-rotating perceptron. 

\section{Experimentation, results and future work}

\begin{figure}[h]
  \centering
    \includegraphics[width=0.99\textwidth]{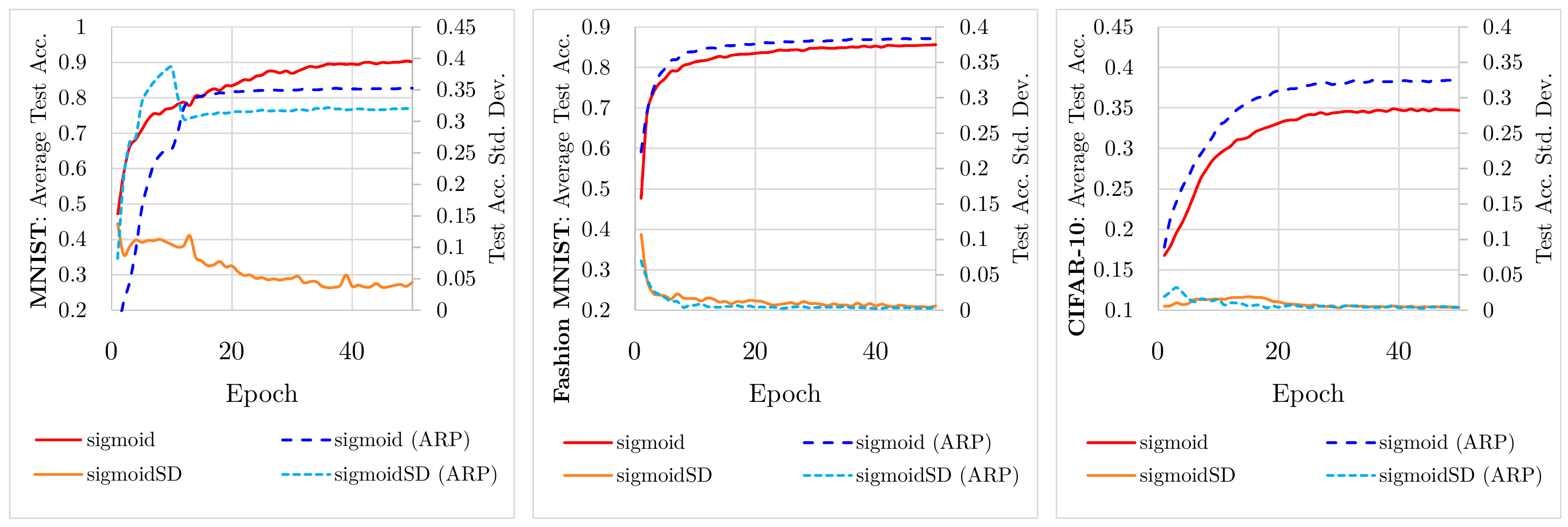}
  \caption{Comparison of the test prediction accuracy. Datasets used: MNIST \cite{mnist_lecun1998} (left), Fashion MNIST \cite{fashion_xiao2017} (middle) and CIFAR-10 \cite{cifar_krizhevsky2009} (right). MLP architecture: (input image size)-50-50-40-30-30-20-10. ARP units only in the hidden layers. Number of executions for each dataset: 30. Epochs per iteration: 50. Same initial weights and biases. Batch size: 64. Optimizer: Adam ($lr=0.003$).}
  \label{fig:graficaONOFF}
\end{figure}

We evaluate the effectivity of the ARP unit, with the unipolar sigmoid activation function, in an MLP architecture using three benchmark datasets (MNIST, Fashion MNIST and CIFAR-10).  The hyperparameters used were: $L=4$ (because the sigmoid $\sigma(z)$ is not saturated for $z \leq |4|$) and $\mathbf{x}_Q\in \mathbb{R}^{n}$ with all its components being $1.1$ (because input data is scaled to the range $|x|\leq 1 < x_Q = 1.1$). Figure \ref{fig:graficaONOFF} shows the test prediction accuracy and its corresponding standard deviation (SD). We observe a notorious improvement of test accuracy in CIFAR-10 with ARP units with respect to classic units. In Fashion MNIST, the improvement is also noticeable but at lower extend. In MNIST no improvement in accuracy is observed, but a high variability of results were found.  
These results show a potential of ARP to deal with the VGP, though further experimentation is needed to confirm  this trend and to understand the behavior of the units with different activation functions, hyperparameter configurations, optimizers and network architectures.

\newpage

\subsubsection*{Acknowledgments}

The authors would like to thank Diego Ugarte  La Torre for his helpful comments about the manuscript.


\bibliographystyle{IEEEtran}
\small 
\bibliography{IEEEfull,mybibfile}

\end{document}